\documentclass[10pt,twocolumn,letterpaper]{article}

\usepackage{cvpr}
\usepackage{times}
\usepackage{epsfig}
\usepackage{graphicx}
\usepackage{amsmath}
\usepackage{amssymb}
\usepackage{booktabs}
\usepackage{caption} 
\usepackage{float}
\usepackage{array}
\usepackage{makecell}
\usepackage{tabularx}
\usepackage{multirow}
\usepackage{verbatimbox}
\usepackage{algpseudocode}
\usepackage{algorithm}
\usepackage{array}
\usepackage{multirow}
\usepackage{dsfont}
\newcolumntype{Y}{>{\centering\arraybackslash}X}

\usepackage[pagebackref=true,breaklinks=true,letterpaper=true,colorlinks,bookmarks=false]{hyperref}

\cvprfinalcopy 

\def\cvprPaperID{5345} 

\ifcvprfinal\fi
\begin{document}

\title{Car Pose in Context: Accurate Pose Estimation with Ground Plane Constraints}

\author{Pengfei Li\textsuperscript{1}\thanks{ The
first three authors contributed equally.}, Weichao
Qiu\textsuperscript{1}\footnotemark[1], Michael Peven\textsuperscript{1}\footnotemark[1], Gregory D. Hager\textsuperscript{1}, Alan L. Yuille\textsuperscript{1}\\
\textsuperscript{1}Johns Hopkins
University\\
{\tt\small \{lipengfeizju, qiuwch, mpeven\}@gmail.com, hager@cs.jhu.edu, alan.l.yuille@gmail.com}}
\maketitle

\begin{abstract}

Scene context is a powerful constraint on the geometry of objects within the scene in cases, such as surveillance, where the camera geometry is unknown and image quality may be poor. In this paper, we describe a method for estimating the pose of cars in a scene jointly with the ground plane that supports them. We formulate this as a joint optimization that accounts for varying car shape using a statistical atlas, and which simultaneously computes geometry and internal camera parameters. We demonstrate that this method produces significant improvements for car pose estimation, and we show that the resulting 3D geometry, when computed over a video sequence, makes it possible to improve on state of the art classification of car behavior. We also show that introducing the planar constraint allows us to estimate camera focal length in a reliable manner.



\end{abstract}



\newcommand\norm[1]{\left\lVert#1\right\rVert}



\section{Introduction}

Crowded urban environments present a challenge for reliable six degree of freedom (6-DoF) pose estimation causing problems from scale and occlusion. On the other hand, the density of information that makes this problem difficult can be exploited to improve results. Here, we present a method that takes global scene context into account and jointly estimates the ground plane with the 6-DoF position of each detected car. With the key assumption that cars lie on the ground plane, and the normal of a car and ground plane should be parallel, our bottom-up and top-down method can perform self-calibration when camera parameters are missing, predict a ground plane, and achieve our goal of estimating 6-DoF car pose - all at the same time. 

The procedure we present here does not rely on the Manhattan World assumption \cite{coughlan2001manhattan}, in which obvious vertical and horizontal lines are present. This allows it to be applied on a larger variety of scenes and backgrounds. The constraint from the ground plane helps to eliminate outliers, removing the noise from individual car estimations that have issues with scale and motion-blur. Furthermore, it provides vital context information for small and blurry cars. This is a generic framework which can be integrated with any keypoint-based object pose estimation algorithm and can be extended beyond cars to arbitrary object categories.

\begin{figure*}[ht!] 
\centering
\includegraphics[width=0.72\linewidth]{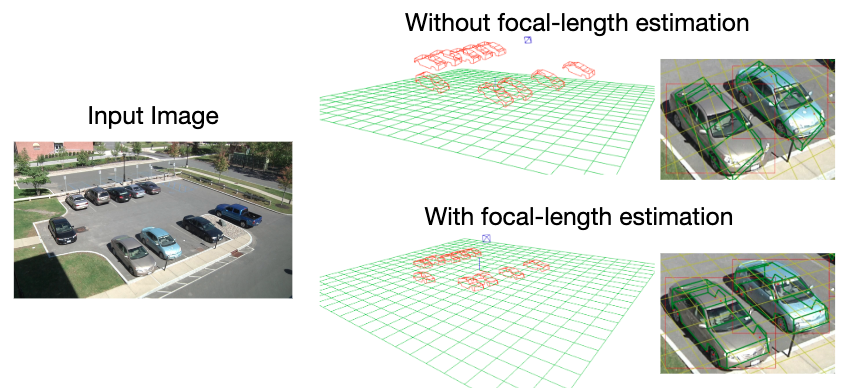}
\vspace{-0.3cm}
\caption{Effect of focal length estimation on the ground-plane and 3D rotation of the cars} 
\label{fig:focal_length}
\end{figure*}

We qualitatively evaluate various components of our method in an ablative manner, for example in Figure \ref{fig:focal_length} we can see that focal length estimation is important for the correct rotation of the ground plane, and without it the cars have unrealistic orientation. We also quantitatively evaluate the pose estimation results from this method, outperforming other models on a synthetic dataset that we created, and which we publicly release with this paper. Furthermore, we demonstrate that our method improves performance over a baseline on real data - the publicly available KITTI Dataset. Additionally, we evaluate the use of our method's accurate pose estimations on down-stream tasks, showing improvement over two baselines on the vehicle-centric actions in an activity recognition dataset. Lastly, this method also enables copying car behavior in the real world to a synthetic world in order to generate large amounts of synthetic data for autonomous driving and car activity classification. In contrast to simulation based behavior~\cite{Dosovitskiy17}, copying real to synthetic enables behavior realism from large-amounts of unannotated videos.

Our contribution can be summarized as follows. First, we propose an optimization scheme to jointly solve camera calibration and pose estimation. Second, we experimentally demonstrate that using our accurate focal length and ground plane estimations can significantly improve performance of 6-DoF pose estimation across multiple datasets. Furthermore, we show how our reliable pose estimates can be used on down-stream tasks, improving activity recognition results over state of the art. Lastly, we publicly release a synthetic parking lot dataset with ground-truth annotations for evaluating 6-DoF car pose.

\section{Related Work}
\textbf{Pose Estimation}: 6-DoF pose estimation has wide applications from virtual reality and robotic manipulation to autonomous driving. An intuitive solution for pose estimation is to compare an image with templates of different object poses, and select the most possible pose. Classical methods including \cite{gu2010discriminative, hinterstoisser2011gradient} use hand-designed gradient templates, which have difficulties in regards to scalability. Recent works \cite{wohlhart2015learning} use end-to-end models to extract object descriptors for different objects and views, however, these template-based methods see performance degradation when estimating unseen objects or views.

Another solution is to train a neural network to directly output the object pose \cite{massa2016crafting, Su_2015_ICCV, sundermeyer2018implicit}. When formulating pose estimation as a regression problem, combining auto-encoders with neural networks are a good choice \cite{sundermeyer2018implicit} as pure neural networks have issues dealing with the non-linearity in rotation.
Some researchers \cite{kehl2017ssd, ChiLiUni} formulate rotation as a classification problem, by discretizing object rotation into a series of bins. Assuming class independence can help prevent outputs from regressing to the mean for symmetric objects. SSD6D \cite{kehl2017ssd} estimates 6-DoF pose by classifying discrete bins of Euler angles, then estimating object translations by fitting projections into 2D bounding boxes. Bin-delta networks such as MVn-MCN \cite{ChiLiUni} propose a pose discretization in SE(3) space, in which the network can output a coarser-grained pose bin and a small offset to the bin. However, most of these end-to-end methods are not transferable between different scales and objects.

More recent methods, including SOTA method \cite{peng2019pvnet} in 6-DoF pose estimation, adopts a two-stage method. They first localize object keypoints in the image robustly, then optimize objects pose to best fit into these keypoints. Motivated by the success of human keypoint localization, several methods \cite{Krishna_ICRA2017, oberweger2018making, pavlakos20176} localize semantic keypoints with a stacked-hourglass \cite{newell2016stacked} model. Although these methods output 6-DoF object pose, camera intrinsics are still needed for the optimization process. 

Compared to existing methods, our method can estimate pose without knowing camera focal length and can be generalized to unseen objects within the same category.


\textbf{Camera Self-calibration}: Unlike camera calibration, camera self-calibration means finding the camera's intrinsic parameters without capturing manually created objects like a calibration checkerboard \cite{zhang2000flexible}. Classical methods \cite{caprile1990using, kanatani2005statistical, schindler2004expectation, wildenauer2012robust} extract line segments and cluster them into orthogonal candidates for camera calibration. Recent methods \cite{Brouwers2016AutomaticCO, liu2011surveillance} exploit pedestrian 2D direction or tracking results in images to estimate vanishing points to aid with the calibration process. These methods tend to work well in complex man-made environments, making use of the Manhattan World assumption \cite{coughlan2001manhattan}, encountering failure cases in scenes that are empty or have a cluttered background.

Another option to solve these problems is using deep neural networks to directly regress the camera's intrinsic parameters. As explained by \cite{Hartley2001MultipleVG}, camera intrinsics can be calculated directly from horizontal lines and vertical vanishing points. \cite{HoldGeoffroy2017APM, Workman2016HorizonLI} use deep neural networks to estimate the horizon line and the network is robust in natural image and generalizable between different scenes. \cite{Abbas2019AGA} use an end-to-end model to regress both horizontal line and vertical vanishing point. However, the outputs of deep neural network are less explainable compared to classical calibration methods. 

\textbf{Surface Normal Estimation}:
Estimating surface normals has applications in human robot interaction for service robotics and adding virtual objects on a plane in AR. Previous works \cite{himmelsbach2010fast, McDaniel2010GroundPI} need expansive RGB-D camera or point clouds from LiDAR measurements, which constrains their application scenarios.
Recent works including \cite{Eigen2014PredictingDS, Hoffman2016LearningWS, Man2019GroundNetMG} can estimate surface normal with monocular images in a multitask network and \cite{zou2018layoutnet} can even be used to recover the geometry of a whole indoor scene. However, an end-to-end model is affected by dataset statistics and may not be able to transfer between different scales and scenarios.
Our method relies on the pose of objects on a plane to vote for the ground plane, which is applicable across different objects and scenes.

\textbf{Scene Context to Improve Local Estimation}: Context information has been considered as prior information for many classical works~\cite{tu2008auto}. The reconstruction of a 3D scene has been proven to be effective for many tasks, such as 3D human pose estimation~\cite{PROX:2019} or scene understanding~\cite{zhang2014panocontext}. 

\textbf{Copying Real Behavior to a Virtual World}:
Modern deep neural networks require a lot of data for training. The data acquisition and labeling is expansive and time-consuming. In addition, the human labeling process incorporates varying degrees of subjectiveness and inconsistency. Generating synthetic images is helpful but there exists a big domain gap between real images and synthetic ones. Virtual KITTI \cite{Gaidon2016VirtualWorldsAP} can copy real images in KITTI dataset to synthetic ones to maintain the content; however, they require accurate human annotations for the objects. \cite{Kar2019MetaSimLT} can generate statistically appropriate synthetic images for downstream tasks without human annotations. However, when solving tasks like car activity analysis, which requires object behavior realism between frames, this method may fail. Our method can copy real to synthetic without manual annotation, but still keeps the behavior realistic.

\section{Problem Formulation}

\subsection{Pose Estimation without Plane Constraint}

Given 2D keypoint locations, a naive method to estimate object pose is to apply Perspective-n-Point (PnP) algorithms \cite{lepetit2009epnp}. The goal is to minimize the loss function shown in Equation~\ref{naive_method}, where $\pi$ is the projection function given by camera parameters (e.g. focal length, distortion factors). $R_k$, $T_k$ is the pose of object $k$,  and $X_{ki}$ is the $i^{th}$ keypoint on object $k$. $x_i^k$ is the position of $i^{th}$ keypoint in pixel level. 


\begin{equation}
    {\setstretch{1.5}
    \begin{array}{crl}
         \underset{R,T}{\min} &  E_{repoj}^k &= \sum_{i=1}^n \| {\pi(R_k X_{ki} + T_k) - x_i^k} \| ^2 \\ 
         s.t. & I &= R_k^T R_k\\
    \end{array}}
    \label{naive_method}
\end{equation}
However, since the focal length is unknown to us, this method cannot be applied directly. There are methods to solve PnP problem with unknown focal length~\cite{DLSPnP, PnPf}, but they have two assumptions: 1) they require accurate keypoint estimation, 2) they are only applicable on rigid objects. These assumptions are invalid in our real world scenario. First, it is difficult to get accurate keypoints due to occlusion, low-resolution, and motion-blur. Furthermore, cars can be seen in a variety of shapes.

In order to handle the shape deformation, we formulate our problem as keypoint estimation for deformable object, rather than rigid object keypoint estimation~\cite{PnPf,Lepetit2008EPnPAA}. 

$$ X = \bar X + \sum_{j = 1}^m \lambda_{j}V_{j} $$

where $n$ keypoints of the object are denoted by $X$, $X \in \mathbb{R}^{3\times n}$.



From 3D keypoint annotations of different kinds of cars~\cite{li2017deep}, $\bar X \in \mathbb{R}^{3\times n}$ is their average keypoint position. $V_j \in \mathbb{R}^{3\times n}$ is the $j^{th}$ possible shape variability computed by Principal Component Analysis (PCA). For efficiency, we take the first $m$ components of PCA where $m=6$ in this case. 

Given the detected keypoint positions $x^k \in \mathbb{R}^{2 \times n}$ in 2D image plane, for each object $k$, the goal is to estimate its rotation matrix $R_k \in \mathbb{R}^{3 \times 3}$, translation $T_k \in \mathbb{R}^3$ and the shape deformation coefficient $\lambda_k \in \mathbb{R}^{m}$. Then we use full perspective model and formulate the inference process as an optimization problem.

\begin{equation}
    {\setstretch{1.5}
    \begin{array}{crl}
         \underset{R,T,\lambda, f}{\min} &  E_{repoj}^k &= \sum_{i=1}^n s_i \| {\pi(R_k X_{ki} + T_k) - x_i^k} \| ^2 \\
         & &\quad + \mu \sum_{j = 1}^m \norm{\lambda_{k,j}}^2\\ 
         & X_k &= \bar X + \sum_{j = 1}^m \lambda_{k,j}V_{k,j}\\
         s.t. & I &= R_k^T R_k\\
         & \lambda_{k,j} &\in [0,U_j]\\ 
    \end{array}}
    \label{opt_prob}
\end{equation}

In this equation, $s_i$ is the confidence score of each keypoint provided by a convolutional neural network. The Tikhonov regularizer $\mu \sum_{j = 1}^m \norm{\lambda_{k,j}}^2$ is introduced to penalize large deviations from the mean shape, and $U_j$ is the upper bound of the deformation coefficient. The formulation described in Equation~\ref{opt_prob} is continuous and can be solved with gradient based methods. We can alternately update each variables and fix others. For instance, we can first solve the optimal $R_k$ given the fixed $T_k$ and $\lambda_k$, and then in reverse. However, since the whole problem is not convex, the gradient descent is only applicable locally and an approximate initialization is necessary. A simple method is to remove the deformable terms and solve this with classical PnP algorithms. The closed-form solution is efficient and good enough for initialization.

\subsection{Joint Formulation with Global Context}
When camera focal length is unknown, it is difficult to directly solve Equation~\ref{opt_prob}. First, accurate focal length is fundamental for object pose estimation, because there are ambiguities between object pose and camera focal length. Second, estimating pose for small or occluded cars is difficult. Small noise at the pixel level will make the car pose change dramatically in 3D space. In order to solve these two problems, adding global context constraints is necessary. 

Regressing focal length with a single object is not feasible. In the image space, object scale is determined by camera focal length in addition to its distance. There may exist a series of combinations of focal length, that produces identical looking images at various distances, especially when object is far away compared to its scale. In some cases, these small differences are totally corrupted by the noise in keypoint localization. Therefore, purely optimizing the reprojection error in Equation~\ref{opt_prob} may be affected by random noise, causing the results to be untrustworthy. 


Nevertheless, using global information will help estimate focal length and eliminate ambiguity in single objects. Among several feasible combinations for focal length and object distance, most of them will not make the object pose inconsistent with global context. For instance, incorrect focal length may cause some objects to float in the air or move beneath ground plane. Thus, in our method, the global context assumption is that all cars park in a same flat ground plane. To measure the difference between single object and global context, we use $D_{p2p}^k$ , $D_{v2v}^k$ to measure straight-line distance and angle distance.

\begin{equation}
    D_{p2p}^k=\frac{v_a t_x^k + v_b t_y^k + v_c t_z^k + v_d}{\sqrt{v_a^2 + v_b^2 + v_c^2}}
\end{equation}
\begin{equation}
    D_{v2v}^k =   \arcsin(\frac{(v_a,v_b,v_c)}{\norm{(v_a,v_b,v_c)} } \times \frac{(n_x^k,n_y^k,n_z^k)}{\norm{(n_x^k,n_y^k,n_z^k)}} )\\
    \label{dis_eq}
\end{equation}
In Equation \ref{dis_eq}, suppose object $k$ locates at $(t_x^k, t_y^k, t_z^k)$, with the vertical axis of the object aligns with $(n_x^k, n_y^k, n_z^k)$. Ground plane is represented by $v_a\cdot x + v_b\cdot y + v_c\cdot z + v_d = 0$. We denote the shortest distance from object $k$ to ground plane as $D_{p2p}^k$. We denote angle between an object's vertical axis and the plane norm vector as $D_{v2v}^k$ 

Suppose we have $N$ instances in an image, the overall loss function can be defined as
\begin{equation}
    {\setstretch{1.5}
    \begin{array}{crl}
         \underset{R,T,\lambda,v_{p}, f}{\min} &  E_{loss} &= \sum_{k=1}^N E_{plane}^k + \sum_{k=1}^N E_{repoj}^k\\ 
          & E_{plane}^k &= \mu_1 \norm{D_{p2p}^k}^2 + \mu_2 \norm{ D_{v2v}^k} ^2\\
          &E_{repoj}^k &= \sum_{i=1}^n s_i \| {\pi(R_k X_{ki} + T_k) - x_i^k} \| ^2 \\
         & &\quad + \mu \sum_{j = 1}^m \norm{\lambda_{k,j}}^2\\ 
         & X_k &= \bar X + \sum_{j = 1}^m \lambda_{k,j}V_{k,j}\\
        
         s.t. & I &= R_k^T R_k\\
         & \lambda_{k,j} &\in [0,U_j]\\ 
         
    \end{array}}
    \label{opt_prob_overall}
\end{equation}

\begin{figure*}[htp!]
\centering
\includegraphics[width=\linewidth]{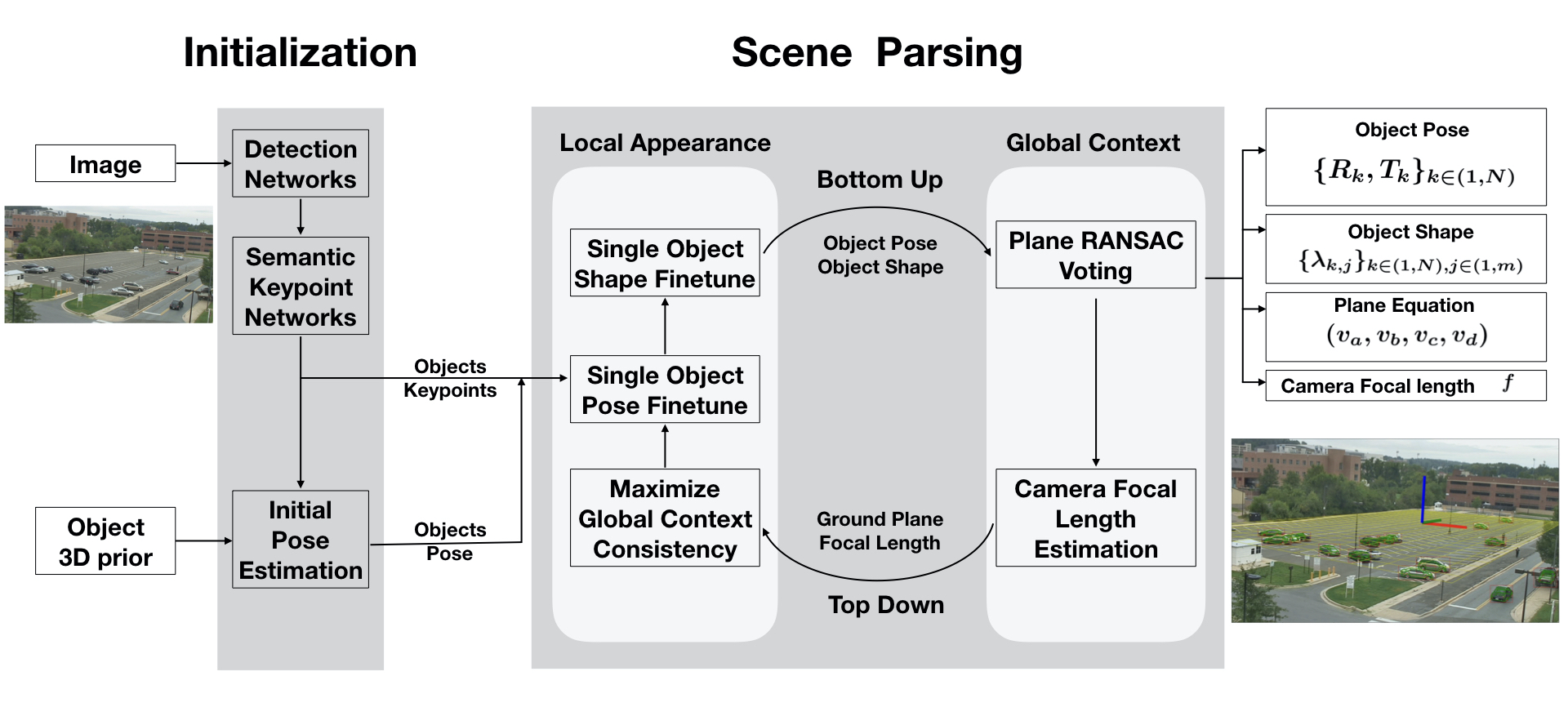}
\caption{General Workflow of World Context Parsing \label{workflow_plane} }
\end{figure*}
In this way, the optimal solution for loss function in Equation~\ref{opt_prob_overall} will also satisfy the global context constraint. The global consistency constraint will make the focal length estimation more accurate and robust to noise. In next section, we will introduce how to extract global information and optimize with it.

\subsection{Joint Optimization}
Now the only remaining problem is how to solve Equation~\ref{opt_prob_overall}. A straightforward way is just to apply gradient descent based methods to optimize this loss function. However, without knowing focal length and plane parameters in advance, we find this method is hard to converge. 

We think there are two reasons to explain this phenomenon. 1) Global context is not given, an initial estimation may not be accurate. Inaccurate global context won't help the optimization, possibly even having a negative effect on convergence. 2) Noise is harmful for gradient descent based methods. A small number of outliers could have large effect on the optimization direction, therefore they should be identified and removed. Our following algorithm can address these two problems.

\begin{algorithm}
    \caption{Joint Optimization Algorithm}\label{alg} 
    \hspace*{\algorithmicindent} \textbf{Input:} Semantic Vehicle Keypoints $\{x^k\}_{k \in (1,N)}$\\
    \hspace*{\algorithmicindent} \hspace*{\algorithmicindent} Object Shape Statistical Atlas $\bar{X}$
    , $\{V_j\}_{j\in (1,m)}$ \\
    \hspace*{\algorithmicindent} \textbf{Output:} Object Pose $ \{ R_k, T_k \}_{k \in (1,N)} $ ; \\
    \hspace*{\algorithmicindent} \hspace*{\algorithmicindent} Object Shape $\{ \lambda_{k,j} \}_{k \in (1,N), j \in (1,m)}$; \\
    \hspace*{\algorithmicindent} \hspace*{\algorithmicindent} Camera Focal Length $f$;\\
    \hspace*{\algorithmicindent} \hspace*{\algorithmicindent} Plane Parameters $( v_a, v_b, v_c, v_d )$
    \begin{algorithmic}[1]
        \State Randomly Initialize $f$, $\{V_j\}_{j\in (1,m)}$, $( v_a, v_b, v_c, v_d )$
        \State Initialize $ \{ R_k, T_k \}_{k \in (1,N)} $ from Direct Linear Transformation
        \State Initialize global context constraint$\mu_1 \leftarrow 0, \mu_2 \leftarrow 0$
        \State Update $ \{ R_k, T_k \}_{k \in (1,N)} $ by Minimizing Loss function in Equation \ref{opt_prob_overall}
        \State $t \leftarrow 0$
        \While{$t < \textrm{MAX\_ITERS}$ and not converged}
            \State $( v_a^1, v_b^1, v_c^1, v_d^1 )$ $\leftarrow $ RANSAC on $\{T_k \}_{k \in (1,N)}$ 
            \State $( v_a^2, v_b^2, v_c^2, v_d^2 )$ $\leftarrow $ RANSAC on $\{R_k \}_{k \in (1,N)}$ 
            \State Normalize $( v_a^1, v_b^1, v_c^1, v_d^1 )$, $( v_a^2, v_b^2, v_c^2, v_d^2 )$
            \State $f \leftarrow v_c^2/v_c^1 \cdot f$
            \State Update $ \{ R_k, T_k \}_{k \in (1,N)} $ by minimizing Equation \ref{opt_prob_overall}
            \State Finetune $\{\lambda_{k,j}\}_{k \in (1,N), j \in (1,m)}$ by minimizing Equation \ref{opt_prob_overall}
            \State Increase global context constraint weights $\mu_1, \mu_2$ 
            \State $t \leftarrow t+1$
        \EndWhile
        
    \end{algorithmic}
    \end{algorithm}

The first step of our method is to get rough estimate of pose. Given randomly initialized camera focal length and ground plane, we find the car pose and shape parameters by minimizing the function in Equation~\ref{opt_prob_overall}. Note that the initial estimate of car pose is not accurate because the camera's focal length is unknown.

The second step to is estimate the global context, namely the ground plane parameter. Since the objects are on the plane, we can fit a plane jointly to the object's 3D position using RANSAC. Their rotation can also be used to help estimate the plane by ensuring the normals are parallel. However, with inaccurate focal length, these two results may differ. We save these two plane results together for the next step.

Next, we refine the camera focal length and object pose. Inconsistency between single object orientation and plane normals comes from two sources. First, noise in keypoint localization will make the pose estimation inaccurate, which incorrectly moves the object away from the ground plane. Second, inaccurate focal length makes the position of the object incorrect even if we locate keypoints perfectly.
Only the first kind of noise is a true outlier and should be removed. Fortunately, the second inconsistency can be solved with correct focal length. Our solution is to estimate the correct focal length with the two plane estimation. Next, this information is used to adjust plane parameter and individual object poses. More details are provided in Section~\ref{sec:optimality}.

Given the new plane parameter and focal length, we can then re-estimate vehicle pose by enforcing the consistency with the new global context. When repeating the above process, we have to balance the weight between global context constraints and the reprojection error dynamically. 
During the optimization process, the pose becomes increasingly more accurate each iteration. This causes the plane estimation result to become more reliable, therefore, we should increase the weight when considering global context consistency. On the other hand, the randomly initialized plane is not reliable at first, then global context weight should be small. This technique will decrease the negative effect of inaccurate global context.


To sum up, the general workflow is show as Figure~\ref{workflow_plane}. We use off-the-shelf object detection and keypoint localization networks to obtain each object's keypoint locations in 2D space, which is used as input for the optimization pipeline. During the optimization process, we use a bottom-up and top-down technique to enforce the consistency between local appearance and global context. Finally we can obtain camera focal length, object poses in 3D space, and ground plane parameters.

\subsection{Optimality of Focal Length and Ground Plane Refinement\label{sec:optimality}}

As we introduced in the previous section, inconsistency between single objects and global context come from two sources and we want to split this into two parts and deal with them differently. 

We expand the reprojection error in Equation~\ref{opt_prob}. For each object, the reprojection error can be written as Equation~\ref{weak_eq}. 




\begin{equation}
    \underset{R_o^{cam}, T_o^{cam}, f}{\arg \min} \sum_j^{N_{kpt}} \norm{s\cdot \begin{bmatrix}
    f  & 0 & u_0 \\
    0  & f & v_0 \\
    0  & 0 & 1 
    \end{bmatrix}\cdot \begin{bmatrix}
    x_j^{cam}\\
    y_j^{cam}\\
    z_{j}^{cam}
    \end{bmatrix}
    - p^{img}_{j}}^2
    \label{weak_eq}
\end{equation}
For the $j^{th}$ keypoint on an object, its position in camera coordinates is denoted as $(x_j^{cam}, y_j^{cam}, z_j^{cam})$, which is determined by the pose of this object $R^{cam}_{o}$, $T^{cam}_{o}$.

The full perspective model is approximated through a weak perspective projection model. We make this choice because in most surveillance videos, the object is far away from the camera in comparison to its scale. In the weak model, each keypoint's depth $z_j^{cam}$ is approximated by the depth of the average keypoint $z_{avg}^{cam}$. Suppose the object center is the arithmetic mean of a object's keypoints, then $z_{avg}^{cam}$ equals the translation of the object along the $z$-axis, denoted as $z_o^{cam}$. 

Given the camera focal length $f$, the optimal solution of Equation \ref{weak_eq} is  $T_{opt}$,$R_{opt}$. From the optimal solution, we suppose the ground truth of plane equation is $Ax_j^{cam} + By_j^{cam} + Cz_j^{cam} + D = 0$.

Now suppose our initial estimate of focal length is not accurate and scaled by factor $\lambda_s$, the optimization becomes
\begin{equation}
    \begingroup
    \renewcommand{\arraystretch}{1.2}
    \setlength\arraycolsep{2pt}
    \underset{R_o^{cam}, T_o^{cam}, f}{\arg \min}\sum_j^{N_{kpt}} \norm{s
    \kern-1pt\cdot\kern-3pt
    \begin{bmatrix}
    \lambda_s \kern-3pt\cdot\kern-3pt f  & 0 & u_0 \\
    0  & \lambda_s \kern-3pt\cdot\kern-3pt f & v_0 \\
    0  & 0 & 1 
    \end{bmatrix}
    \kern-3pt\cdot\kern-3pt
    \begin{bmatrix}
    x_j^{cam}\\
    y_j^{cam}\\
    z_o^{cam}
    \end{bmatrix}
    - p^{img}_{j}}^2
    \label{weak_err}
    \endgroup
\end{equation}
To expand matrix multiplication in Equation~\ref{weak_err}, we can have
\begin{align*}
    x_j^{img} &= \frac{\lambda_s \cdot f\cdot x_j^{cam} + u_0 \cdot z_o^{cam}}{z_o^{cam}}\\
    y_j^{img} &= \frac{\lambda_s \cdot f\cdot y_j^{cam} + v_0 \cdot z_o^{cam}}{z_o^{cam}}\\
\end{align*}
The optimal solution of Equation~\ref{weak_err} is denoted as $'T_{opt}$,$'R_{opt}$. A trivial solution is 
$'R_{opt} = R_{opt}$
\begin{equation}
    'T_{opt} = \begin{bmatrix}
    'x_j^{cam}\\
    'y_j^{cam}\\
    'z_o^{cam}
    \end{bmatrix} = \begin{bmatrix}
    x_j^{cam}\\
    y_j^{cam}\\
    \lambda_s \cdot z_o^{cam}
    \end{bmatrix}
\end{equation}
This optimal solution has same reprojection error as the ground truth. Now from the inaccurate focal length, the equation now becomes $A\cdot {'}x_j^{cam} + B\cdot {'}y_j^{cam} + \frac{C}{\lambda_s}\cdot {'}z_j^{cam} + D = 0$.  
\newline From these properties, we have three findings
\begin{enumerate}
    \item Without the ground plane constraint, the problem is an ill-posed problem. Equation~\ref{opt_prob} has infinite number of optimal solutions. 
    \item There exists a way to adjust object pose estimation without sacrifice the reprojection error.
    \item Inaccurate focal length initialization will make the ground plane normal inconsistent with object normals. The inconsistency only occurs in $Z$ axis and we can compare them to correct the focal length.
\end{enumerate}

\subsection{Activity Recognition}\label{activity_recognition}

The temporal stream of 3D positions for objects such as cars can be examined under the lens of activity recognition in video. Events such as turns and changes in acceleration, can be classified into various actions. To model these types of behaviors, we use recurrent neural networks to predict activities. Specifically, given the track (bounding box at each frame) of an object in an RGB video, we wish to classify the activity at each frame during the track. This is known as action recognition, namely, the joint segmentation and classification of activities in a temporal sequence.

Formally, this can defined as the prediction of action label $Y_t \in \{0,1\}^C$, where $C$ is the number of classes and $Y_t$ is the one-hot encoding of the action class at time $t$ for $1 \leq t \leq T$ of an object tracked for $T$ frames. This is modeled as $Y_t = f(P_{t} - P_{t-1})$, where $P_t$ is the 3D position of an object at frame $t$, and $f$ is a bi-directional LSTM taking the velocity of the object as input. The LSTM is a natural fit for this type of problem, as we want to model the long-term dynamics of an object's temporal behavior.

\section{Experiments}

In the following experiments, our object keypoint input is given by the network proposed by Chi et. al\cite{chili3D}. We choose this network as it is shown to be robust to occlusion and truncation. In Equation \ref{opt_prob_overall}, with our alternating minimization scheme, in every step, the loss of each object is decoupled. This means the Jacobian is very sparse, which helps to speed up our optimization.  
Because the baseline methods \cite{pavlakos20176,pavlakos20176} require known camera intrinsics, in the following pose estimation evaluation experiments, the focal length is given as input.  

\subsection{Synthetic 6-DoF Pose Estimation}\label{exp:synthetic}
\textbf{Pose Estimation}
We adopt the ADD evaluation metrics proposed in \cite{hinterstoisser2012model}. 
This evaluation metric is based on calculating the point-wise average distance in 3D space. 
The ground truth model's point positions are generated from its ground truth object pose and CAD model. 
The estimated point position is the output of our algorithm and other baseline methods. 
We use this to calculate the average 3D distance between the estimated points and their ground truth correspondence. 

For baseline methods outputs the position of several keypoints, we pick the corresponding keypoint on the object and evaluate average point distance. To make the distance more intuitive, we normalize by the object's diameter. 


\begin{table}[h!]
\renewcommand{\arraystretch}{1.1}
\centering


\begin{tabularx}{\linewidth}{@{\hskip 1pt}c@{\hskip 4pt} Y@{\hskip 4pt} Y@{\hskip 4pt} Y@{\hskip 4pt} Y@{\hskip 4pt} Y@{\hskip 1pt}}
& \multicolumn{5}{c}{Threshold (Normalized distance)} \\
\cline{2-6}
\addlinespace[3pt]
\rule{0pt}{9pt} \multirow{1}{*}{Method} &  0.4 & 0.8 & 1.2 & 1.6 & 2.0\\
\midrule
Georgios et al.\cite{pavlakos20176} & 0.00 & 0.00 & 0.00 & 0.00 & 0.00\\
\addlinespace[3pt]
Krishna et al.\cite{Krishna_ICRA2017} & 53.47 &  83.12 & 90.35 & 92.14 & 93.05\\
\addlinespace[3pt]
\makecell{Our Method \\\addlinespace[-3pt]\textit{w/o plane-constraint}}    & 66.32 & 87.91 & 94.62 & 96.91 & 97.71\\
\addlinespace[3pt]
\makecell{Our Method \\\addlinespace[-3pt]\textit{with plane-constraint}}   & \textbf{77.99} & \textbf{93.40} & \textbf{96.86} & \textbf{97.84} & \textbf{98.30}\\

\bottomrule 
\end{tabularx}
\caption{Pose estimation accuracy with the ADD metric in our synthetic dataset. The distance threshold is normalized by object diameter. Details on the failure of Georgios et al.\cite{pavlakos20176} is given in Section \ref{exp:synthetic}}
\label{tab:synthetic_pose}

\end{table}
For Georgios et al.\cite{pavlakos20176} method, we directly use the open-source code. Their method performs well on PASCAl 3D+\cite{xiang_wacv14}. In this dataset, objects are relatively big and there are few occlusions. However, in our synthetic data, there are heavy occlusions in the parking lot and cars in the image can be small.

\textbf{Viewpoint Estimation}
Another commonly used metric to evaluate viewpoint estimation is the Average Orientation Precision (AOP) metric proposed in \cite{xiang_wacv14}. This metric mainly focuses on viewpoint correctness. For the original AOP criterion, the detection output is considered correct if the 3D bounding box overlap ratio is within 70\% and the error of estimated viewpoints is less than a threshold. However, calculating arbitrary 3D bounding box overlap is non-trivial, so we only evaluate the viewpoint precision of baseline methods and our method. Similar to the evaluation process in \cite{Krishna_ICRA2017}, we use different thresholds for viewpoint precision evaluation.

\begin{table}[h!]
\renewcommand{\arraystretch}{1.1}
\centering


\begin{tabularx}{\linewidth}{@{\hskip 1pt}c@{\hskip 4pt} Y@{\hskip 4pt} Y@{\hskip 4pt} Y@{\hskip 4pt} Y@{\hskip 4pt} Y@{\hskip 1pt}}
& \multicolumn{5}{c}{Threshold (Radians)} \\
\cline{2-6}
\addlinespace[3pt]
\rule{0pt}{9pt} \multirow{1}{*}{Method} &  0.14 & 0.21 & 0.28 & 0.35 & 0.42\\
\midrule
Georgios et al.\cite{pavlakos20176} & 4.55 & 9.10 & 13.87 & 18.15 & 21.86\\
\addlinespace[3pt]
Krishna et al.\cite{Krishna_ICRA2017} & 84.06 &  88.33 & 90.07 & 91.10 & 91.78\\
\addlinespace[3pt]
\makecell{Our Method }   & \textbf{88.34} & \textbf{94.55} & \textbf{96.34} & \textbf{97.22} & \textbf{97.59}\\

\bottomrule 
\end{tabularx}
\caption{Object viewpoint estimation performance on our synthetic dataset.}
\label{tab:viewpoint}

\end{table}


\subsection{Real-World 6-DoF Pose Estimation}
We also evaluate our algorithm in KITTI\cite{Geiger2012CVPR}, which is a large-scale and realistic evaluation benchmarks for autonomous driving. The 3D object benchmark provides vehicle pose and bounding box annotation in 3D space. When the images have more than three vehicles in view, our global context constraint is feasible. Otherwise, our method is almost the same as other baseline method. To validate the effectiveness of plane constraints, we only evaluate on images with more than three vehicles in view. The results are shown in Table \ref{tab:KITTI_pose}. We also perform experiments without plane constraints, in which the algorithm performs almost the same. We find that the error mainly occurs in depth axis, so the inaccurate pose is still on the plane. Plane constraint is not able to correct this type of error.

\begin{table}[h!]
\renewcommand{\arraystretch}{1.3}
\centering

\begin{tabularx}{\linewidth}{@{\hskip 2pt}cYYYYY}
& \multicolumn{5}{c}{Threshold (Normalized distance)} \\
\cline{2-6}
\addlinespace[3pt]
\rule{0pt}{9pt} \multirow{1}{*}{Methods} &  0.4 & 0.8 & 1.2 & 1.6 & 2.0\\
\midrule 
Georgios et al.\cite{pavlakos20176} & 0.04 & 0.16 & 0.20 & 0.34 & 0.49\\
Krishna et al.\cite{Krishna_ICRA2017}      & 6.43 & 12.66 & 17.49 & 23.85 & 27.75 \\
Our Method    & \textbf{12.29} & \textbf{22.64} & \textbf{28.78} & \textbf{36.81} & \textbf{41.84}\\

\bottomrule 
\end{tabularx}
\caption{Pose estimation accuracy with ADD metric in the KITTI dataset. The distance threshold is normalized by object diameter.}
\label{tab:KITTI_pose}
\end{table}

\subsection{Copying Car Behavior To Synthetic Video}
Simulating real world car behavior is an important task for building realistic car simulator. Thanks to the advantage of computer graphics, achieving visual realism is becoming much easier. But there are still a lot of challenges in generating realism car behaviors. Previous works such as CARLA\cite{Dosovitskiy17} generate synthetic videos with manually designed car behaviors. The car behavior on the road depends on many factors, such as road condition, pedestrian behavior, etc. This makes it really difficult to design a simple algorithm to simulate real world car behavior.

\begin{figure}[htp!]
    \centering
    \includegraphics[width=\linewidth]{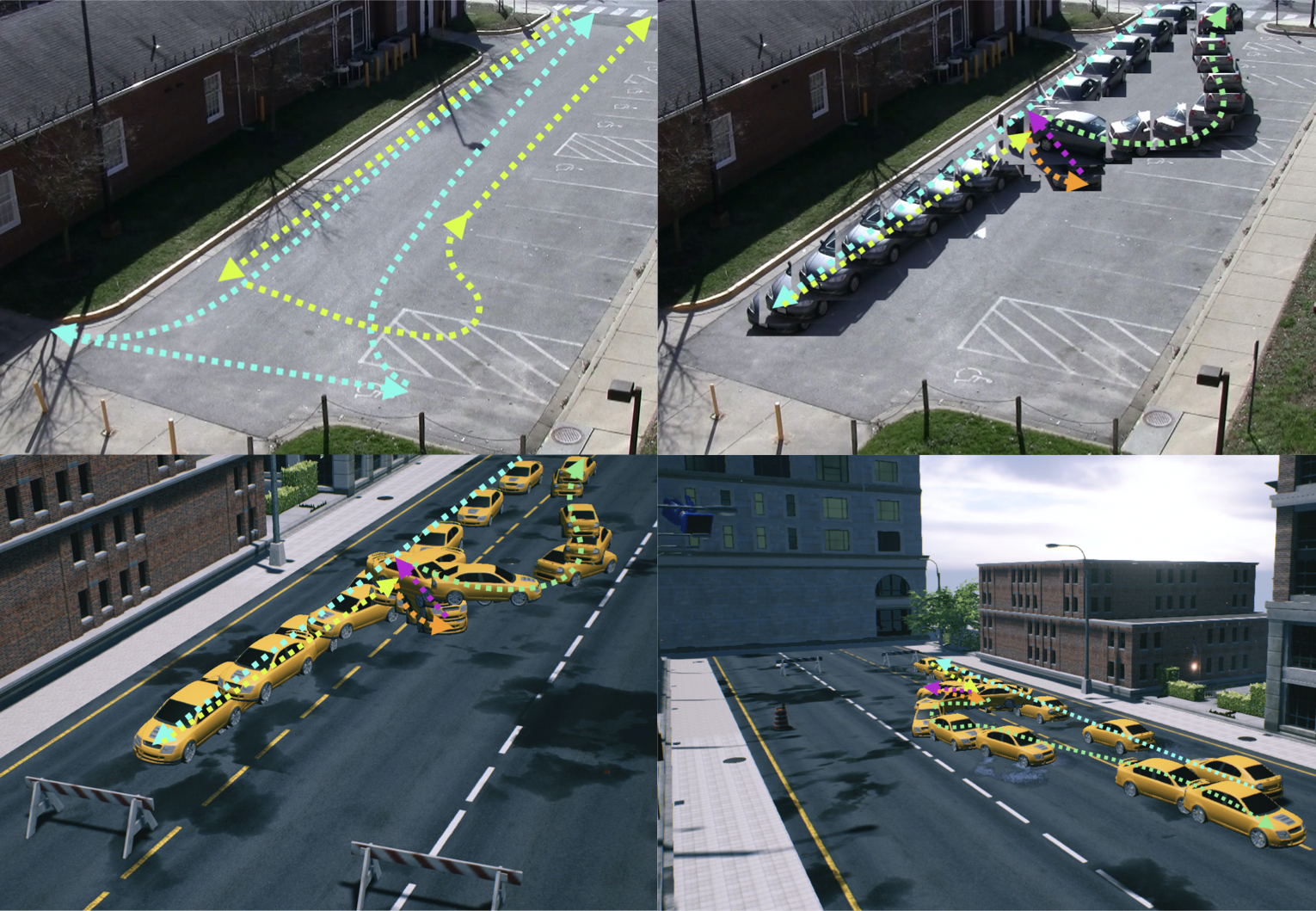}
    \caption{Copy real car behaviors into the synthetic world. \textbf{Upper-left} is some normal car trajectories. \textbf{Upper-right} is an example of abnormal car behavior. This car first go straight, then drive reversely, then go straight, then drive reversely and turn, finally it turns left and go straight. \textbf{Bottom-left} is the copied behavior in UnrealCV\cite{qiu2017unrealcv}, which copies the same camera intrinsic, extrinsic and vehicles poses. \textbf{Bottom-right} is the video captured in another position and viewpoint}
    \label{fig:copy_real}
\end{figure}

In order to solve this challenge, one feasible solution is copying real world car behavior into the virtual world. There are millions of street surveillance cameras on the road, if an algorithm can copy real-world car behaviors into a simulator. Then it is possible to simulate different views and conditions for autonomous driving or other purposes. This can potentially have a huge impact. Fig. \ref{fig:copy_real} shows an example of copying a video clip from VIRAT Video Dataset \cite{oh2011large}. In this video, the behavior of the car is abnormal and hard to manually design. 

\begin{figure*}[htp!]
    \centering
    \includegraphics[width=\linewidth]{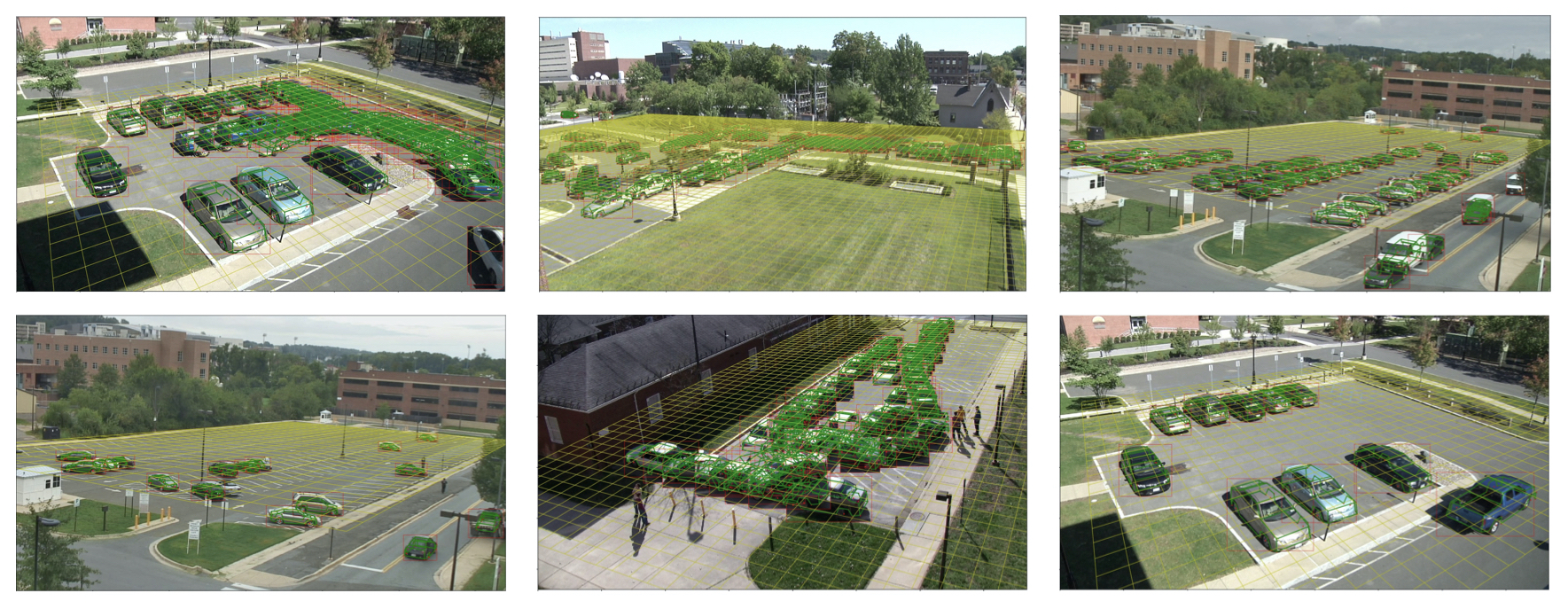}
    \vspace{-20pt}
    \caption{Visualizations of estimated ground planes and car poses produced from our full pipeline}
    \label{fig:vis_ground_plane}
\end{figure*}

\subsection{Activity Recognition}
To evaluate the effectiveness of pose estimation on down-stream tasks, we look at performance of using our 3D pose estimates for activity recognition on the VIRAT Video Dataset \cite{oh2011large}. This is a surveillance dataset consisting of multiple views of different parking lots. Both car bounding boxes and car activities are annotated on a per-frame basis. The list of annotated activities involving cars are: \textit{Turning Right}, \textit{Turning Left}, \textit{U-Turn}, \textit{Stopping}, \textit{Starting}, and \textit{Moving}. Additionally, there is a background class, which can be labeled as \textit{Not Moving}.

For experimental evaluation, we use the outputs of our method into the bi-directional LSTM model described in Section \ref{activity_recognition} to produce per-frame activity predictions. Specifically, the input is velocity of our estimated 3D car location in the ground plane, $(dx_{3D}, dy_{3D})$, assuming we can ignore the height of car above the ground plane for the these activities. As a baseline comparison, we use the car's 2D location in the image plane as input, specifically, the velocity of the center pixel in the bounding box. We use the same LSTM architecture, keeping all hyperparameters consistent. Furthermore, we compare our geometric-based approach to a video-based approach. The trimmed video clips are trained and evaluated with the I3D \cite{carreira2017quo, gleason2019proposal} network, the current state of the art activity recognition network on this dataset.

\renewcommand{\arraystretch}{1.3}
\begin{table}[h!]
\centering
\begin{tabular}{@{\hskip 2pt}c@{\hskip 4pt} @{\hskip 4pt}c@{\hskip 4pt} @{\hskip 4pt}c@{\hskip 4pt} @{\hskip 4pt}c@{\hskip 4pt} @{\hskip 4pt}c@{\hskip 2pt}}
\toprule
Method & Input & Accuracy & Precision & Recall \\
\midrule
2D & image position & 97.5\% & 0.48 & 0.66 \\
I3D \cite{carreira2017quo} & video frames & 99.2\% & 0.51 & 0.37 \\
3D & world position & \textbf{99.4}\% & \textbf{0.65} & \textbf{0.75} \\
\bottomrule 
\end{tabular}
\caption{Results averaged across all 7 activities in the VIRAT dataset for the models tested.}
\label{tab:activity_results}
\vspace{-10pt}
\end{table}

We present the results of these experiments in Table \ref{tab:activity_results}. The very high accuracies are due to the number of frames with motionless vehicles. This creates a large class imbalance weighted towards \textit{Vehicle Not Moving}, which is accounted for during training by using a weighted cross-entropy during training for all three of the models; not surprisingly however, they still perform well on parked cars. To get a better sense of the performance, we also include precision and recall, averaged across all classes. These numbers are more telling, demonstrating that the results from our method significantly outperforms the same exact experimental set up but with image-plane velocities as input rather than ground-plane velocities. Furthermore, we also improve over the state of the art deep activity recognition model.

\section{Conclusion}

Global scene context can be exploited to help estimate 6-DoF object pose. We design a method to use this global information by jointly estimating the pose of all cars in a scene with each other, along with the ground plane that supports them. We demonstrate both qualitatively and quantitatively how our method produces accurate car pose, ground plane, and camera parameter estimates. We further show that the 3D car positions produced can be used in a straightforward manner to improve over state of the art methods for activity recognition. In addition to this paper, we publicly release a synthetic parking lot dataset for evaluation of 6-DoF car pose estimation.

\vspace{1.5em}
\textbf{Acknowledgements} Supported by the Intelligence Advanced Research Projects Activity
(IARPA) via Department of Interior/ Interior Business Center (DOI/IBC)
contract number D17PC00342. The U.S. Government is authorized to
reproduce and distribute reprints for Governmental purposes
notwithstanding any copyright annotation thereon. Disclaimer: The
views and conclusions contained herein are those of the authors and
should not be interpreted as necessarily representing the official
policies or endorsements, either expressed or implied, of IARPA,
DOI/IBC, or the U.S. Government.

{\small
\bibliographystyle{ieee_fullname}
\bibliography{egbib}
}

\end{document}